\documentclass[runningheads,a4paper]{llncs}

\usepackage{hyperref}
\hypersetup{
  pdftitle={Multi-rendezvous Spacecraft Trajectory Optimization with Beam P-ACO},
  pdfauthor={Lu{\'i}s F. Sim{\~o}es, Dario Izzo, Evert Haasdijk, A. E. Eiben},
  pdfkeywords={Beam Search, Ant Colony Optimization, P-ACO, Bilevel optimization, Multi-objective optimization, Spacecraft trajectories},
  bookmarksopen=true,
  bookmarksopenlevel=1,
  colorlinks=true,
  linkcolor=black, citecolor=black,
  urlcolor=blue,
}

\usepackage{makeidx}  % allows for indexgeneration

\usepackage{geometry}
\usepackage[pdftex]{graphicx}
\DeclareGraphicsExtensions{.pdf}

\usepackage[caption=false,font=footnotesize]{subfig}

\usepackage{amsmath}

\usepackage{url}

\makeatletter
\def\blfootnote{\xdef\@thefnmark{}\@footnotetext}
\makeatother

\newcommand{\PACO}{\mbox{P-ACO} }

\urldef{\mailsa}\path|{luis.simoes, e.haasdijk, a.e.eiben}@vu.nl, dario.izzo@esa.int|
\newcommand{\keywords}[1]{\par\addvspace\baselineskip
\noindent\keywordname\enspace\ignorespaces#1}

\begin{document}

%\frontmatter
%\mainmatter

%\pagestyle{headings}  % switches on printing of running heads
%\addtocmark{Hamiltonian Mechanics} % additional mark in the TOC
%

% ===== TITLE
%
\title{Multi-rendezvous Spacecraft Trajectory\\ Optimization with Beam P-ACO}
\titlerunning{Multi-rendezvous Spacecraft Trajectory Optimization with Beam P-ACO}

% ===== AUTHORS
%
\author{Lu{\'i}s F. Sim{\~o}es\inst{1} \and Dario Izzo\inst{2} \and Evert Haasdijk\inst{1} \and A. E. Eiben\inst{1}}
\authorrunning{L.F. Sim{\~o}es et al.}

\institute{Vrije Universiteit Amsterdam, The Netherlands\\
\and European Space Agency, ESTEC, The Netherlands
\mailsa%\\
%\mailsb\\
%\mailsc\\
}

\maketitle

\begin{abstract}
The design of spacecraft trajectories for missions visiting multiple celestial bodies is here framed as a multi-objective bilevel optimization problem.
A comparative study is performed to assess the performance of different Beam Search algorithms at tackling the combinatorial problem of finding the ideal sequence of bodies.
Special focus is placed on the development of a new hybridization between Beam Search and the Population-based Ant Colony Optimization algorithm.
An experimental evaluation shows all algorithms achieving exceptional performance on a hard benchmark problem.
It is found that a properly tuned deterministic Beam Search always outperforms the remaining variants.
Beam P-ACO, however, demonstrates lower parameter sensitivity, while offering superior worst-case performance. Being an anytime algorithm, it is then found to be the preferable choice for certain practical applications.

\keywords{Beam Search, Ant Colony Optimization, P-ACO, Bilevel optimization, Multi-objective optimization, Spacecraft trajectories.}

\end{abstract}

\section{Introduction}
\label{sec:intro}

\blfootnote{Code available at \url{https://github.com/lfsimoes/beam_paco__gtoc5}.}
\setcounter{footnote}{0}

The design of multi-rendezvous spacecraft trajectories poses a considerable challenge to aerospace engineers. This is due, in part, to the combinatorial nature of the problem that emerges with the increase in number of bodies to visit along a mission (e.g., planets, moons, asteroids).
The complexity of the task stems from an interplay of multiple factors under optimization, including: the decision of which of the bodies of interest to visit, the order in which they are to be visited, and the design of the actual trajectory arcs to take the spacecraft between them.
Maximization of such a mission's scientific return may demand for as many bodies to be visited as possible, in the shortest possible amount of time, while consuming the lowest possible amount of propellant mass.
The underlying optimization problem can be seen as a variant of the well known Traveling Salesman Problem (TSP), with nodes corresponding to the celestial bodies under consideration, and edge weights a function of the costs (time and mass) to propel the spacecraft between them. These costs vary as bodies move in space along their trajectories, but also as a function of the spacecraft's state: a lighter spacecraft that has already shed some of its mass is simpler to maneuver.

As evidence into the complexity, and relevance, of the above described problems, consider the following:
since 2005, the aerospace engineering community has periodically organized GTOC, the Global Trajectory Optimization Competition \cite{gtoc:portal}. In it, different groups have taken turns at creating ``nearly-impossible" problems of interplanetary trajectory design to pose to the community. Of the 8 competitions organized to date, 4 were multiple-asteroid rendezvous problems of the kind considered here, and 3 others were multiple fly-by problems posing similar combinatorial optimization challenges.

In \cite{Izzo:16:gtoc7_chap}, a multi-objective Beam Search algorithm is described, and applied to a low-thrust model of the GTOC7 problem. In this research, we propose a series of extensions to it. First, we provide an improved orbital phasing indicator, and a procedure to create from it a probability distribution over candidate bodies to extend missions with. Second, we hybridize the Beam Search procedure with the well known Ant Colony Optimization algorithm \cite{ACObook}.
We conclude by evaluating the resulting algorithms on a Lambert model of the GTOC5 problem \cite{GTOC5:Grigoriev}.
Two main research questions are investigated in this paper:
\begin{enumerate}
\item Can the randomization of Beam Search, via probabilistic branching choices, improve performance?
\item Does the pheromone-based positive reinforcement of sequences lead to improved performance?
\end{enumerate}

This paper is organized as follows:
in Sec. \ref{sec:related_work} we list related work in the combinatorial optimization of spacecraft trajectories.
Sec. \ref{sec:beam} describes the Beam Search algorithm, and the proposed randomized variants.
In Sec. \ref{sec:gotc5} we describe the GTOC5 problem used in our experiments, and
in Sec. \ref{sec:optimiz_gtoc5} the interfacing between it and the search algorithms.
Sec. \ref{sec:expeval} reports on an experimental evaluation, and
Sec. \ref{sec:discuss} discusses its results.
Conclusions are drawn in Sec. \ref{sec:conclusion}.

\section{Related work}
\label{sec:related_work}

Beam Search \cite{Bisiani:87:BS,wilt2010comparison} has emerged as the \textit{de facto} standard approach to tackle the combinatorial optimization sub-problems present in most GTOC competitions. Though at times called by other names, it is common to find the general architecture of a tree search that has its computational cost bounded via the selection of a limited number of nodes to branch at each depth-level (non-selected nodes at that depth being immediately discarded).
We can find examples of such algorithms in the winning solutions to GTOC4 \cite{grigoriev2013choosing}, GTOC5 \cite{GTOC5:Petropoulos}, and in the second ranked solution to GTOC7 \cite{Izzo:16:gtoc7_chap}, which the present research builds on.
The Lazy Race Tree Search described in \cite{GTOC6:ESA:GrandTour}, which at the time presented the best known solution to the GTOC6 problem, can also be seen as a Beam Search variant. In it, the ``beam'' is composed of all nodes, possibly originating from different tree depths, that fall within a given mission time window. The most promising nodes in that sliding window are branched, and the remaining ones discarded.

Evolutionary Algorithms have been explored as an alternative to solve combinatorial problems in mission analysis. In the GTOC5 problem considered here, for instance, \cite{GTOC5:Izzo} and \cite{Gad:2011:PhD} used Genetic Algorithms with ``hidden genes", to evolve chromosomes encoding asteroid sequences.
These approaches were however outperformed in the GTOC5 competition by tree-based approaches.
In \cite{Izzo:2015:dynTSP} an evolutionary approach is described for designing debris removal missions. In this highly dynamic trajectory problem, the Inver-over Genetic Algorithm was found to provide competitive solutions to those constructed by different approaches.

Ant Colony Optimization (ACO) \cite{ACObook} was used by some teams in the GTOC competitions over the years. However, to our knowledge, no scientific publication has been produced to date with the details of such deployment. The most successful use of ACO in a GTOC competition was possibly that by the NASA/JPL team, winners of GTOC7. According to the GTOC portal \cite{gtoc:portal}, a ``very competitive solution was found by JPL using an Ant Colony Optimization approach. Eventually a different solution turned out to be better and was thus submitted''\footnote{The NASA/JPL team's GTOC7 submission report and workshop slides, containing details of their ACO deployment, can be found in the GTOC portal \cite{gtoc:portal}.}.
Other applications of ACO algorithms to optimize the sequences of bodies to visit along a spacecraft's trajectory can be found in \cite{ceriotti2010mga,ceriotti2010automated,stuart2016design}. In them, the test problems over which algorithms are evaluated have few bodies to select from ($\approx 10$), and the found sequences visit $\approx 5$ bodies. In contrast, sequences of up to 17 asteroids are assembled here, from a database of 7075 available asteroids.

A hybridization of Beam Search and ACO was previously presented in \cite{blum2005beamaco}.
A different hybridization is introduced here, ``Beam P-ACO'', that differs mainly in the ACO variant under use, and in being a multi-objective algorithm.

\section{Beam Search}
\label{sec:beam}

Beam Search \cite{Bisiani:87:BS,wilt2010comparison} is a tree search algorithm where computational cost is % \linebreak
bounded by employing heuristics that allow for non-promising solutions under construction to be discarded. It can be executed as a variant of depth-first or breadth-first search. When operating as a variant of breadth-first search, as is done here, Beam Search traverses the tree one depth-level at a time. From all the solutions generated at one level, only a limited subset (the so called ``beam'') will be selected for carrying over to the next level. An evaluation of solutions' quality determines whether they are included in the beam, or instead permanently discarded from the search. The size of the beam is designated as the ``beam width'', and is here represented as $bw$.
The extension of partial solutions in the beam can be performed towards all possible successor nodes, or instead towards only a limited number, enabling further control over the search's computational cost. A ``branching factor'' parameter, here represented as $bf$, indicates that each solution in the beam will be extended only towards $bf$ successor nodes, chosen as a function of how good the solutions they lead to are estimated to be.

In summary, at one level of the tree, each of the beam's $bw$ solutions will be expanded towards $bf$ new nodes, resulting in $bw * bf$ new partial solutions. These are then evaluated, and the $bw$ best of those will become the beam that is carried over to the next tree level.
The search process is therefore driven by two heuristics: 1) $h_s$, which evaluates partial solutions (a path down from the tree's root node), and 2) $h_e$, which evaluates candidate successor nodes for extending solutions with. The more accurately these heuristics point to the complete optimal solution, the more successful the search will be. Beam Search being an ``incomplete" search algorithm however, the identification of the globally optimal solution is not guaranteed \cite{wilt2010comparison}.

\subsection{Multi-objective heuristics}
\label{sec:mobs}

The first extension we introduce to the conventional Beam Search framework is the addition of multi-objective heuristics.
That is, either $h_s$, or $h_e$ (or both) will evaluate partial solutions or candidate extensions, respectively, according to multiple objectives.
In such a scenario, it is then necessary to employ techniques that will allow for the ranking of alternatives, or probabilistic selection among them, that take into account the multiple evaluations.

In our current implementation, $h_s$ evaluates partial solutions according to multiple objectives, while the evaluation of candidate extensions by $h_e$ remains single-objective.
Beam Search must then be able to select at each tree level the best $bw$ solutions from among the newly generated extensions.
To that end, we employ a Pareto dominance approach. Specifically, we apply non-dominated sorting \cite{Deb:MOEA:book} to the pool of extensions, and include in the beam as many of the best Pareto fronts as needed to reach the beam size $bw$. That process will most likely lead to a final Pareto front whose full inclusion in the beam would exceed $bw$. A tie-breaker criterion must then be defined to determine which of those (equally good) solutions to keep, and which to discard (see Sec. \ref{sec:trajeval}).

\subsection{Probabilistic branching}
\label{sec:stoch_beam}

In the ``Stochastic Beam'' algorithm, the beam's construction remains deterministic, but the branching stage will now be subjected to probabilistic decisions.
Given a solution in the beam, with a probability $q_0$ it will be extended towards the $bf$ nodes with best $h_e$ evaluation. With a probability of $1 - q_0$, the choice of $bf$ nodes will instead be a biased sampling without replacement, proportional to $h_e$.
Note that through a parameter setting of $q_0=1$ the algorithm reverts to the previously described (deterministic) Beam Search.

Beam Search will always converge to the same solution, given the same root node. Stochastic Beam however, can be executed multiple times, and possibly converge to different solutions in each run.
These solutions may outperform those found by Beam Search, by including links to nodes that Beam Search incorrectly prunes, due to ranking above the $bf$ threshold.
Non-determinism therefore provides a degree of robustness against imperfections in the $h_e$ heuristic.

\subsection{Hybridization with Ant Colony Optimization}
\label{sec:beam_paco}

A hybridization of Beam Search with Ant Colony Optimization brings two mains changes to the algorithmic framework: 1) multiple tree searches are now performed, in consecutive runs designated as ``generations", and 2) positive feedback takes place, in the form of ``pheromones" that change the $h_e$ heuristic evaluation of candidate successor nodes, therefore biasing the dynamics of tree searches in subsequent generations.

Of the many ACO variants in existence, we chose to hybridize with the Population-based Ant Colony Optimization algorithm (P-ACO), introduced in \cite{Guntsch2002a,Guntsch2002b,Guntsch2003,Guntsch:PhD}.
A detailed analysis in \cite{IRIDIA:PACO:rep} found it to be ``competitive to the state-of-the-art ACO algorithms with the advantage of finding good solution quality in a shorter computation time''.
Also, a recent thorough benchmarking of a high number of approaches to solve the Traveling Salesman Problem found P-ACO to be the best of the tested global optimization algorithms, as well as the best overall algorithm when seeded and hybridized with local search \cite{weise2014benchmarking}.

In ``Beam P-ACO", a partial solution at node $i$, with available successors $S$ evaluates the quality of extending towards node $j$ by
\[
h_e^{'}(i,j) = \frac{\tau(i,j)^\alpha h_e(i,j)^\beta}{\sum_{s \in S}{\tau(i,s)^\alpha h_e(i,s)^\beta}}
\]
where $\tau$ is the pheromone concentration along an edge, and $h_e$ (known as $\eta$ in the common ACO notation) is the problem-specific heuristic. 
The weighting factors $\alpha$ and $\beta$ determine the relative contributions of pheromone and heuristic values to the branching decision.
As in the previously described Stochastic Beam algorithm, with a probability $q_0$ the solution is extended towards the $bf$ nodes with best $h_e^{'}$ evaluation, while with a probability of $1 - q_0$, that choice is instead a biased sampling without replacement proportional to $h_e^{'}$.
A setting of $\alpha=0$ would result in all pheromone information being ignored, and Beam P-ACO would then revert to the Stochastic Beam algorithm.

The pheromone concentration $\tau$ along an edge takes in \PACO discrete values in a given range $[\tau_{init}, \tau_{max}]$.
We define these parameters as $\tau_{init}=1/(n-1)$ and $\tau_{max}=1$, where $n$ is the total number of nodes. This implements the convention from \cite{Guntsch2002b} of having the row/column sum of initial pheromone values be 1 (assuming a problem where revisits are disallowed, and the diagonal of the pheromone matrix is therefore 0).
Given the pheromone range, and $k$, the maximum size of a population of top solutions, we can define $\tau_\Delta = (\tau_{max} - \tau_{init}) / k$ as the pheromone increment deposited in an edge by a solution in the population that follows it along its path. If $l$ solutions in the population include the edge $(i,j)$, its pheromone concentration will then be $\tau(i,j) = \tau_{init} + l \tau_\Delta$.

\subsection{Pareto elitism}
\label{sec:paretoelite}

To complete the definition of Beam P-ACO, a population update model must be defined, a model that handles multi-objective evaluations produced by the $h_s$ heuristic.
We propose here a variation of the method defined in \cite[Sec. 3.1]{Guntsch2003}.
An archive will collect the set of non-dominated solutions found so far.
At the end of each tree search, the best found solutions are merged into it.
After updating the archive, the population that defines the pheromone matrix is reset.

In standard P-ACO \cite{Guntsch2002a}, the population is implemented as a FIFO-queue (each generation's best solution enters the population, possibly displacing the oldest solution it contains so as not to exceed the population size $k$).
We propose here instead to have one FIFO-queue per each of the problem's $n$ nodes, each with a size limit of $k$.
When resetting the population, all FIFO-queues are emptied. Then, solutions in the archive are shuffled, and one by one, they are added to the population (of edges). Each edge $(i,j)$ in those solutions will result in $j$ being added to the $i$th FIFO-queue. The $i$th FIFO-queue then directly maps to the $i$th row of the pheromone matrix, and its contents define which nodes do pheromones most bias a solution at node $i$ to branch towards.

If solutions visit all $n$ nodes, as is the case in many problems (e.g., TSP), this process will result in only the last $k$ solutions of the shuffled archive adding pheromones to the population.
A random unbiased sampling of $k$ archive solutions would then be sufficient.
However, in problems where solutions include only some of the nodes (such as here in the GTOC5 problem, where $n=7075$, but a solution will visit $<20$ nodes), only small amounts of pheromone would be deposited, and plenty of information in the archive would be ignored.
By following in such problems the process described above, the pheromone matrix may now receive contributions from more than $k$ solutions, while still only receiving $\leq k$ contributions at the level of each individual node. This way, in the limit, all solutions in the archive may end up depositing pheromones.

\section{The GTOC5 trajectory design problem}
\label{sec:gotc5}

The trajectory design problem posed in the 5th edition of the Global Trajectory Optimization Competition (GTOC5) is used as benchmark in the current research.
The full problem specification can be found in \cite{GTOC5:Grigoriev}. The problem dataset, along with additional information related to this edition of the competition, can be found in the GTOC portal \cite{gtoc:portal}.
Our current work makes use of the problem model developed by the competition's 4th ranked team \cite{GTOC5:Izzo}.

In the GTOC5 problem, a spacecraft leaves the Earth at some point along an 11-year time-window, to embark on a 15 year (max.) mission of asteroid exploration. The spacecraft starts with a mass of 4000 kg, of which 3500 are reserved for propellant mass, and the scientific equipment used at asteroids.
A total of 7075 asteroids are available as possible targets to visit along the mission.
The exploration of a single asteroid is carried out in two stages. First, the spacecraft must rendezvous (match position and velocity) with the asteroid, and leave there a 40 kg scientific payload. Later in the mission, the spacecraft performs a fly-by of the asteroid, and sends a 1 kg ``penetrator'' towards it. Upon impact, this penetrator would release a cloud of debris, that would be investigated by the payload left there. Partial scores are given for the rendezvous and fly-by maneuvers. An asteroid on which both are performed contributes 1 point to the full score. In the problem models developed by most teams, including the one used here, the problem is simplified by having an asteroid's fly-by maneuver performed immediately after its rendezvous: the spacecraft departs the asteroid, moves some distance away, and then accelerates back towards it. Under this simplification, a trajectory that manages to complete all pairs of maneuvers on a given sequence of $m$ asteroids will score $m$ points.

GTOC5 was won by a team from NASA/JPL with a score 18 trajectory \cite{GTOC5:Petropoulos}.
In contrast, the model being used here, with the initial trajectory conditions listed in Sec.~\ref{sec:expeval:setup}, has only ever allowed the discovery of trajectories with at most a score 16.
Nevertheless, for the goal of evaluating the performance of algorithms that solve the combinatorial part of the problem (finding the asteroid sequence that enables the greatest possible score), the used model is perfectly suitable.

\section{Bilevel optimization of GTOC5 trajectories}
\label{sec:optimiz_gtoc5}
% [section on changes/details of how the search is carried out over the GTOC5 problem]

The design of GTOC5 trajectories is here tackled as a multi-objective bilevel optimization problem \cite{Sinha:17:Bilevel}.
At an upper level, optimization seeks to identify a good subset of asteroids, and the order in which they are to be visited. At a lower level, each chosen pair of asteroids triggers an optimization of the trajectory leg that takes the spacecraft between them.
This section details how the different Beam Search variants are employed for solving the upper-level combinatorial problem, and how the lower-level process optimizes transfer legs.

\subsection{Orbital phasing indicators as heuristic estimators}
\label{sec:heuristic}
% [defines $h_e$]

The problem of assembling an efficient heuristic able to help select possible asteroid targets is crucial and very difficult at the same time. The ground truth (i.e. the optimal $\Delta V$ cost obtained optimizing the transfer leg) is far too expensive to be be computed for all possible asteroid targets and for all states encountered along the search. A solution to this problem was recently proposed via the development of so-called orbital phasing indicators \cite{Izzo:16:gtoc7_chap}.
These essentially allow to introduce for each epoch $t_s$ a metric over the set of all possible asteroids, a metric that can, in turn, be used to detect asteroid neighborhoods efficiently.
It was shown in \cite{Izzo:16:gtoc7_chap} how the orbital indicator, defined as $d_o(\mathcal A_1, \mathcal A_2, \Delta T) = |\mathbf x_t - \mathbf x_s|$, where
$
\mathbf x = \left[\frac{1}{\Delta T} \mathbf r(t_s) + \mathbf v(t_s), \frac{1}{\Delta T} \mathbf r(t_s)\right]
$
and $\mathbf r(t_s)$ and $\mathbf v(t_s)$ correspond to the asteroid ephemeris,
positively correlates to some extent to the ground truth (i.e. the asteroids that are actually easy to reach via an orbital transfer). It essentially considers a snapshot at $t_s$ of the asteroid population and, using a zero order approximation for the dynamics, predicts what asteroids are the closest in terms of transfer $\Delta V$. As such, it is bound to neglect the known state of the asteroid population at the arrival time $t_t$, which seems as a loss of available information. A simple modification to the orbital indicator, though, allows to account for the final asteroid geometry and thus to improve the overall correlation to the ground truth. In essence, one can consider the orbital indicator backward in time, starting from the arrival asteroid, to get a new indicator (note that asteroid velocities will have to have their sign inverted). The average between the two, i.e. the orbital indicator and the backward orbital indicator, is what we use here and call improved orbital indicator, defined as $d_{o'}(\mathcal A_1, \mathcal A_2, \Delta T) =  |\mathbf x_t - \mathbf x_s|$, where
\[
\mathbf x = \left[ \frac{1}{\Delta T} \mathbf r(t_s) + \mathbf v(t_s), \frac{1}{\Delta T} \mathbf r(t_s),
\frac{1}{\Delta T} \mathbf r(t_t+\Delta T) - \mathbf v(t_t+\Delta T), \frac{1}{\Delta T} \mathbf r(t_t+\Delta T)\right]
\]

For every node $i$ being branched during the search, to which corresponds a trajectory presently at asteroid $\mathcal A_i$, we compute $d_{o'}(\mathcal A_i, \mathcal A_j, \Delta T)$ for all of the problem's $n$ nodes, assuming a reference transfer time of $\Delta T=125$ days. From it we build a probability distribution over successor nodes as $h_e(i,j) = (1 - p(i,j)/n)^\gamma$, where $p(i,j)$ is the rank in $\{0,\ldots,n-1\}$ of $d_{o'}(\mathcal A_i, \mathcal A_j, \Delta T)$ among all estimated costs. This results in a selection probability that decays exponentially with increasing rank, at a rate tuned through $\gamma$.
A setting of $\gamma=50$ is used in this problem. There is then a $\approx 30\%$ chance of branching towards an asteroid ranked among the 50 best, and $\approx 84\%$ among the 250 best.
Finally, the problem disallows revisits to asteroids. So, for any node $i$, $h_e(i,j)=0$ if $j$ was already visited at any previous point.

\subsection{Optimization of transfer legs}
% [defines $\Delta V$, the ground truth to the above-defined $h_e$]

During the tree search, branching a partial solution into a given node triggers an optimization process.
Its end result will be the definition of the transfer leg that allows the spacecraft to rendezvous with the corresponding asteroid.
Adding the new leg extends the mission's trajectory, which can then be reevaluated by the $h_s$ heuristic.
In the approach followed here, only the trajectory's rendezvous legs need to be optimized. The cost estimates for self-fly-by legs are instead approximated by a linear acceleration model \cite[Sec. 3]{GTOC5:Izzo}.

Each transfer will have a duration in a set time window of $[60, 500]$ days.
A grid of 50 evenly spaced values defines the candidate transfer times, $\Delta T$ ($\approx 9$ days separation between grid points).
For each $\Delta T$, we employ PyKEP's \cite{Izzo:12:ICATT}\footnote{\url{http://esa.github.io/pykep/}} multiple revolution Lambert solver \cite{Izzo:RevisLamb,Izzo:16:gtoc7_chap} to design a trajectory arc having that exact duration. If multiple revolution solutions exist for a given $\Delta T$, the one with lowest $\Delta V$ is chosen.
Two constraints are imposed: 1) $\Delta T$ should exceed the parabolic time of flight given by Barker's equation\footnote{Legs failing this check are immediately discarded, saving computation time that would otherwise be spent generating Lambert arcs with excessive $\Delta V$.}, and 2) the leg's maximum acceleration should be $< 90\%$ of the maximum supported by the spacecraft.
% \footnote{The $90\%$ limit on the maximum acceleration is imposed so as to increase the likelihood the leg will in post-processing have a feasible low-thrust conversion.}
These constraints lead many transfers to have no feasible solution, for any $\Delta T$ (the targeted asteroid is simply unreachable).
From all $\Delta T$ points in the grid that do have a feasible solution, the one with lowest $\Delta V$ is chosen to define the new rendezvous leg.
In this approach, the optimization of one leg then equates to finding the solutions to at most 50 Lambert's problems.
Note that this final $\Delta V$ cost is the ground-truth to the $d_{o'}$ indicator described in the previous section, from which $h_e$ is defined.
Note also that we take here a greedy choice of $\Delta T$, imposing in that way a transfer leg upon the combinatorial search problem that might in the long term be sub-optimal with respect to its own goals.

\subsection{Trajectory evaluation, ranking, and selection}
\label{sec:trajeval}
% [defines $h_s$, and how the search algorithms handle its evaluations]

Trajectories are evaluated by $h_s$ with respect to three criteria: 1) the mission's score, 2) the total mass required for propellant, and scientific equipment left at asteroids, and 3) the total time of flight.
The ideal mission will have the greatest possible score, while requiring the lowest possible amounts of mass and time.

As mentioned in Sec. \ref{sec:mobs}, a Pareto dominance approach \cite{Deb:MOEA:book} is used for handling the multiple objectives. However, in this problem mass and time evaluations are only fairly comparable among trajectories that share the same score.
As such, ranking a set of trajectories involves 1) binning trajectories according to their score, and 2) applying non-dominated sorting \cite{Deb:MOEA:book} over the mass and time costs of trajectories within each bin.
Identifying the top trajectories in a given set takes place by iterating through bins in descending order of score, gradually extracting their Pareto fronts.
If only a subset of a Pareto front's trajectories is required, those with lowest mass cost are favored.

In a tree search, this process is applied at each depth-level to construct the beam with the best $bw$ of the newly extended solutions.
Before that, however, $h_s$ is used for pruning nodes corresponding to missions that require $> 3500$ kg, or $> 15$ years.
Should that result in an empty pool of extended solutions, or the pool otherwise be empty because no feasible transfer legs were found, the tree search has then reached its final level and is terminated.
The beam of solutions carried over from the previous tree level will then be its final output.
In Beam P-ACO, this signals the end of a generation.
The contents of that final beam are then merged with the archive of non-dominated solutions found so far.
By applying the previously described ranking process, the archive will always be a Pareto front of trajectories that all share the maximum score reached to date.
Beam P-ACO will at this point refresh pheromones. Note that the combinatorial problem is asymmetric: an edge $(i,j)$ present in a good solution is not predictive of an edge $(j,i)$ being likely to lead to good solutions. As such, an $(i,j)$ edge in an archived solution only results in pheromones along the $(i,j)$ direction.

\section{Experimental evaluation}
\label{sec:expeval}

An experimental evaluation was carried out to assess the performance of the different Beam Search variants, using the GTOC5 problem as a test case.
Performance is here evaluated in terms of multiple criteria.
Primarily, we care about search algorithms that enable the consistent discovery of asteroid sequences having the greatest possible length (score).
Among equally scored missions, we care for the best possible coverage of the Pareto front of mass and time costs required to achieve such score.
Finally, these considerations must be traded-off against the computational cost to obtain such solutions.

\subsection{Setup}
\label{sec:expeval:setup}

We adopt as measure of computational cost the number of trajectory legs optimized throughout the search. This is in practice the main performance bottleneck, especially if instead of a Lambert model one were to use a low-thrust one.
A threshold of 100000 optimized legs was used as stopping criterion.

\begin{figure*}[t]%[!pt]
\centering
\subfloat{\includegraphics[width=.33\textwidth,viewport = 43 31 536 357,clip]{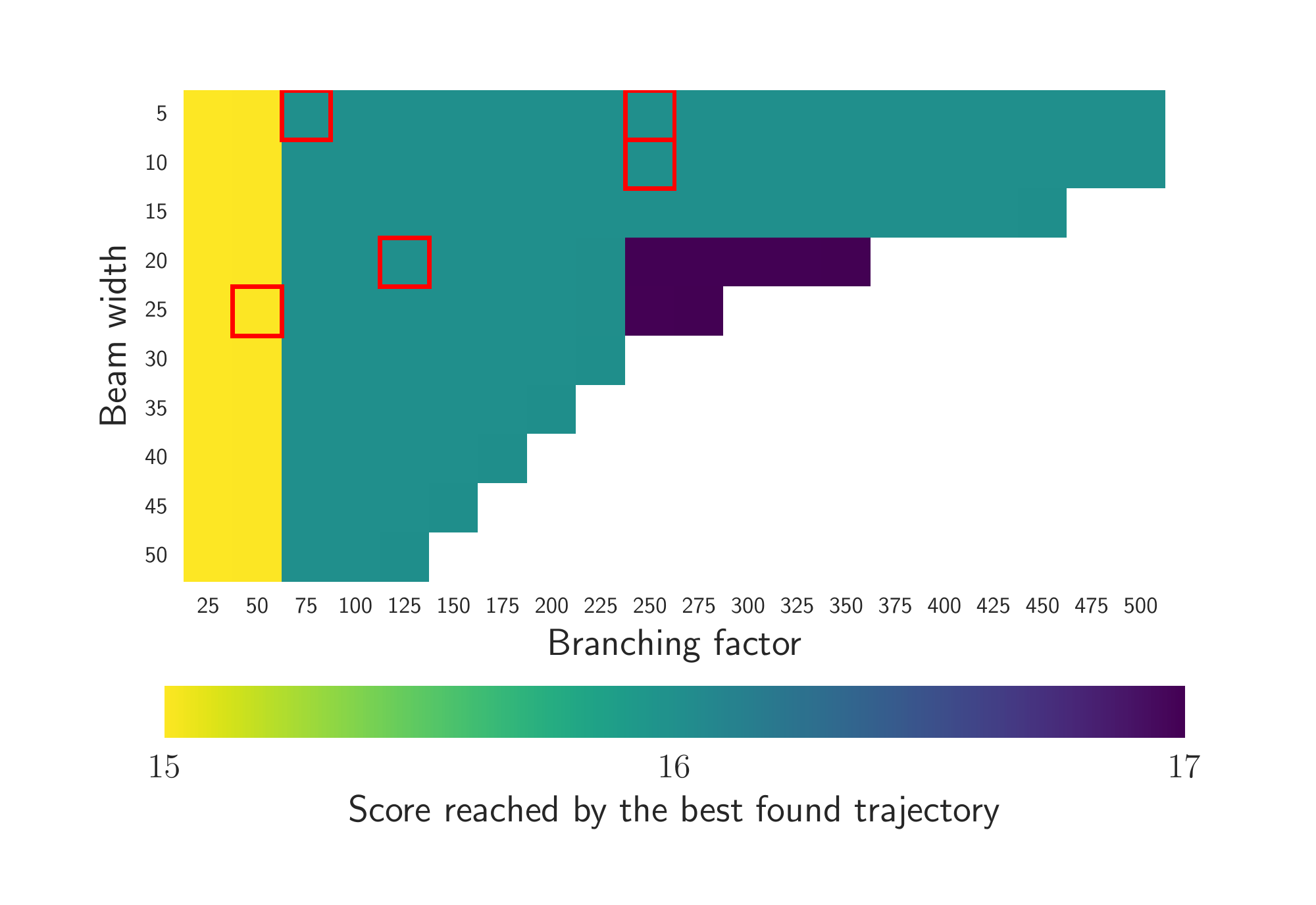}}
\hfil
\subfloat{\includegraphics[width=.33\textwidth,viewport = 43 31 536 357,clip]{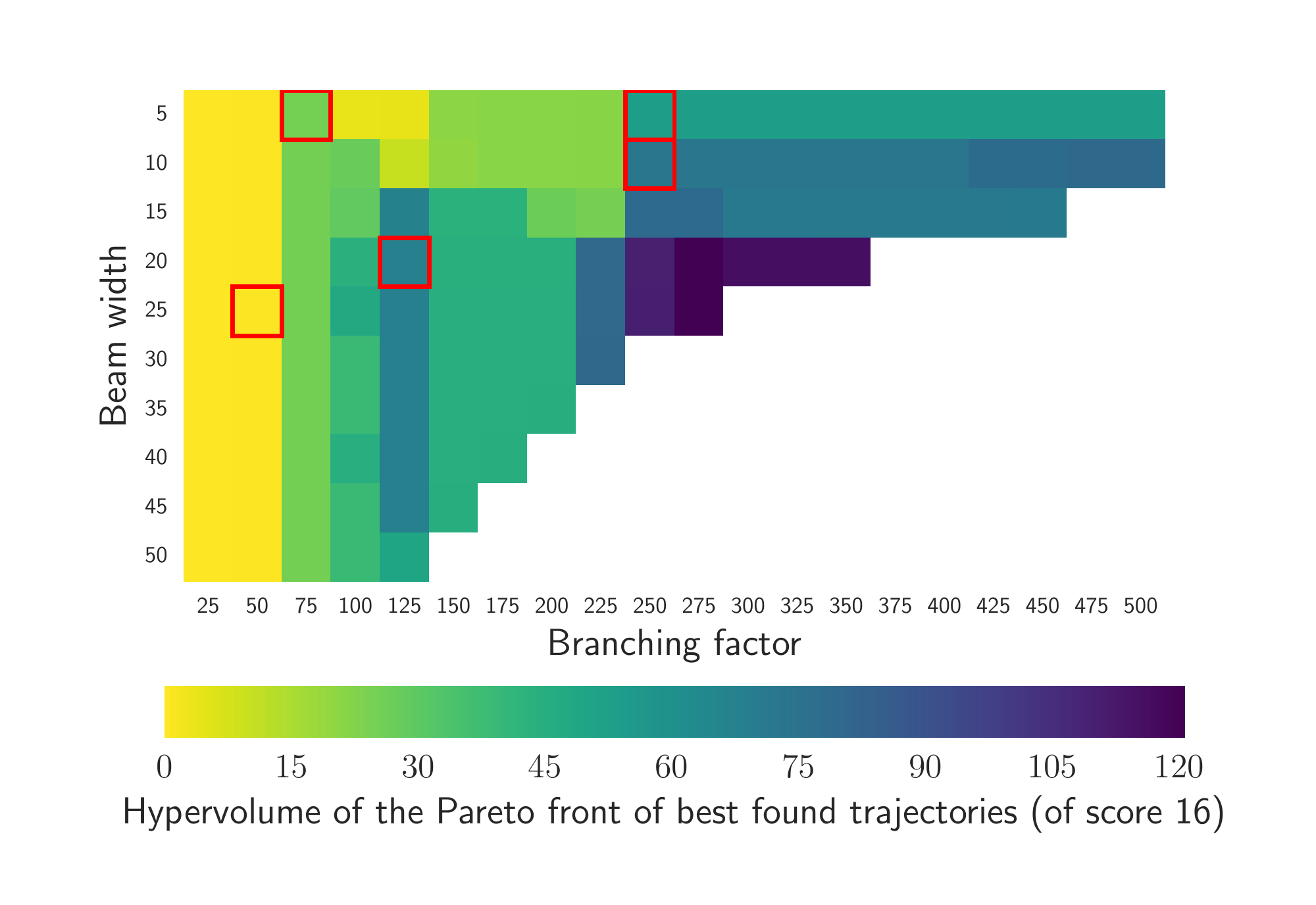}}
\hfil
\subfloat{\includegraphics[width=.33\textwidth,viewport = 43 31 536 357,clip]{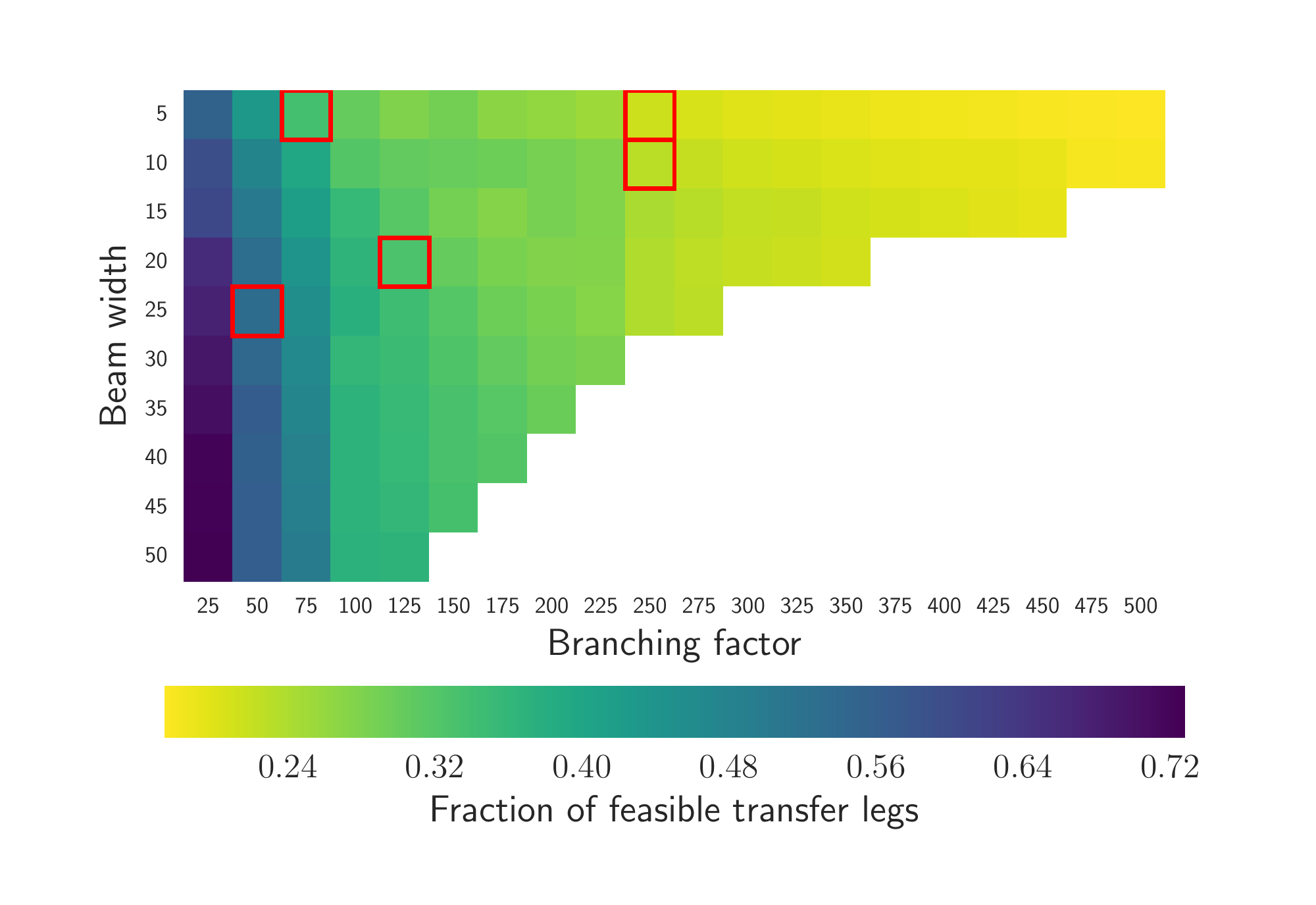}}
\caption{Beam Search results from the parameter sweep over beam width ($bw$) and branching factor ($bf$) configurations (118 setups in total, with costs below the threshold of 100000 optimized trajectory legs). Darker is better. Highlighted in red: configurations used in the Stochastic Beam and Beam P-ACO experiments.}
\label{fig:param_sweep}
\end{figure*}

The performance of deterministic Beam Search was evaluated over a dense grid of settings for the beam width ($bw$) and branching factor ($bf$) parameters. In total, 118 different setups were evaluated, all having upfront an estimated cost of $\leq 100000$ optimized legs required in order to complete execution at tree depth 16, where missions reach and fully score the 16th asteroid (see Fig. \ref{fig:param_sweep}). Being a deterministic algorithm, only a single run was executed per setup.

Stochastic Beam and Beam P-ACO were evaluated under the 5 different configurations of $bf$ and $bw$ highlighted in red in Fig. \ref{fig:param_sweep}. Being stochastic algorithms, 100 independent runs were performed per setup.
Deterministic branching decisions were taken with a probability of $q_0=0.5$ ($q_0=1$ in Beam Search). In Beam P-ACO, pheromone and heuristic values contributed equally to the $h_e^{'}$ heuristic: $\alpha = \beta = 1.0$ (in the other Beam Search variants, implicitly $\alpha=0.0$).
Pheromone concentrations were limited to at most $k=3$ contributions to each node.

All tree searches reported here had the same root node. Its initial conditions were originally obtained during the GTOC5 competition through a time-optimal low-thrust optimization of the launch leg, as described in \cite[Sec. 2]{GTOC5:Izzo} and exemplified in \cite[Sec. 6.1]{Izzo:12:ICATT}. After applying the linear model to define the self-fly-by leg, the initial state is then a trajectory that has already scored 1 point at asteroid \textit{2001 GP2} (id: 1712), and is ready to depart from it at epoch 59325.360 MJD, having already expended 253.518 kg and 198.155 days from the total budgets.

\begin{figure*}[t]%[!pt]
%\vspace{-10pt}
\centering
\subfloat[Beam Search]{\includegraphics[width=.33\textwidth,viewport = 22 8 537 363,clip]{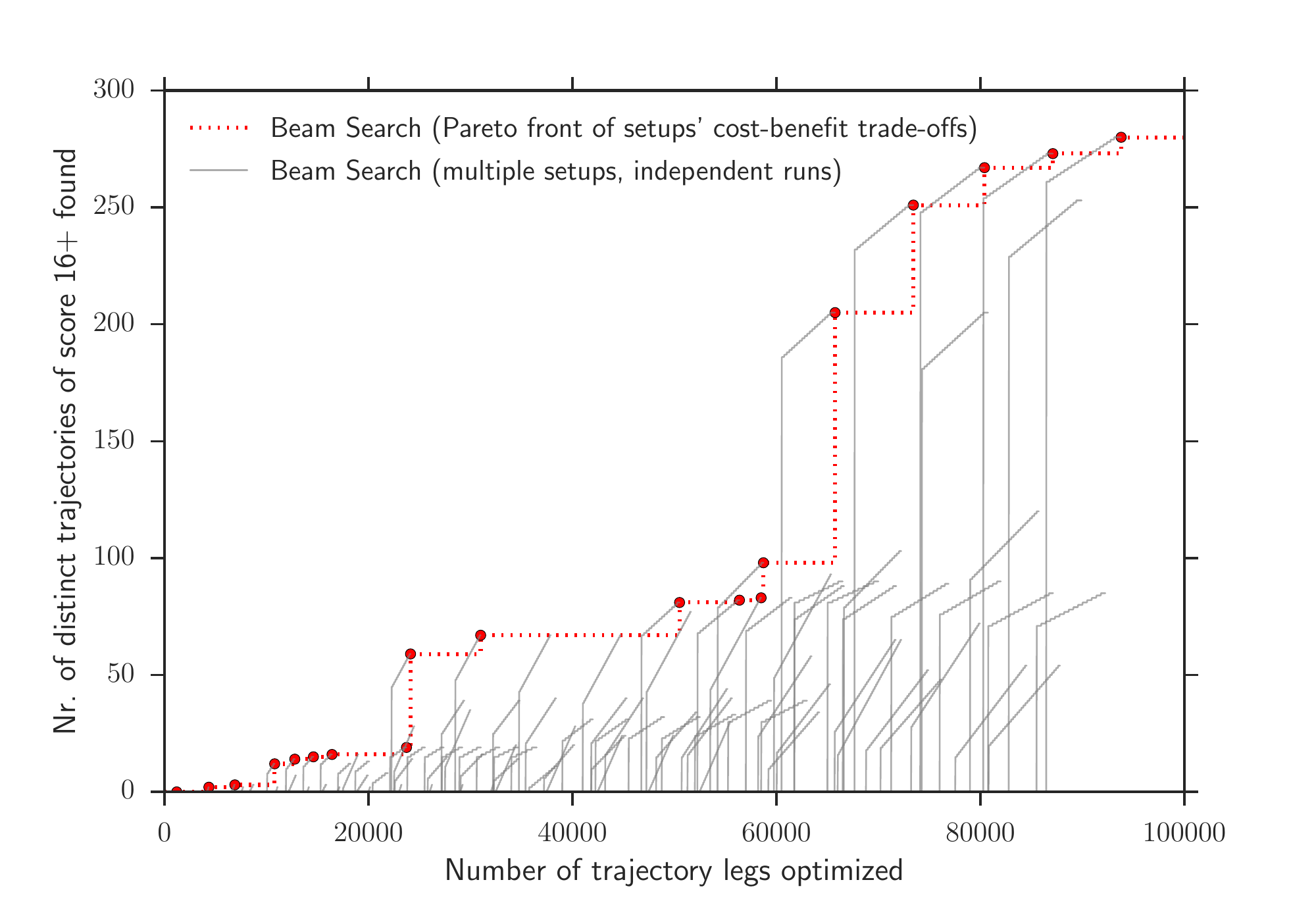}}
\hfil
\subfloat[Stochastic Beam]{\includegraphics[width=.33\textwidth,viewport = 22 8 537 363,clip]{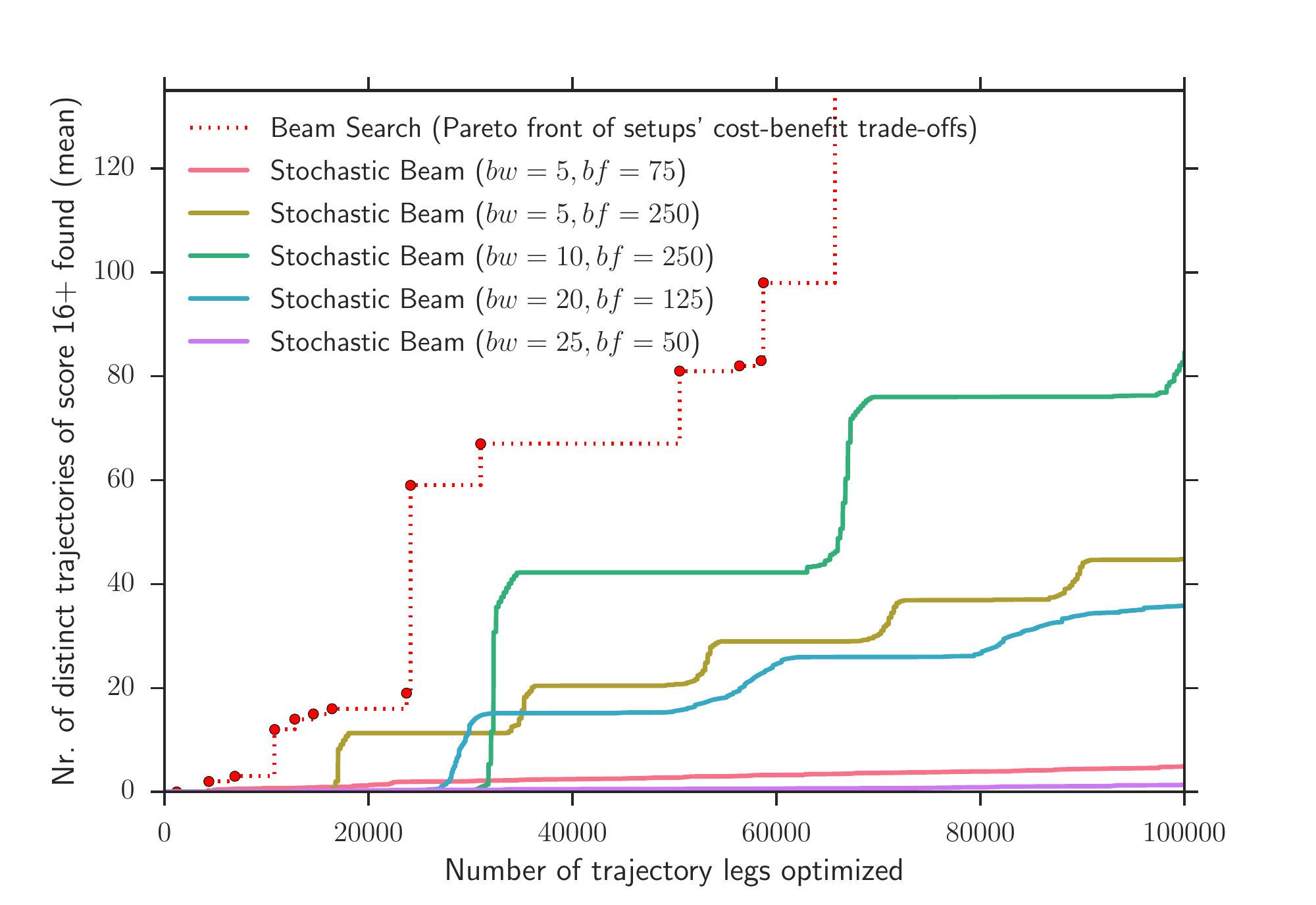}}
\hfil
\subfloat[Beam P-ACO]{\includegraphics[width=.33\textwidth,viewport = 22 8 537 363,clip]{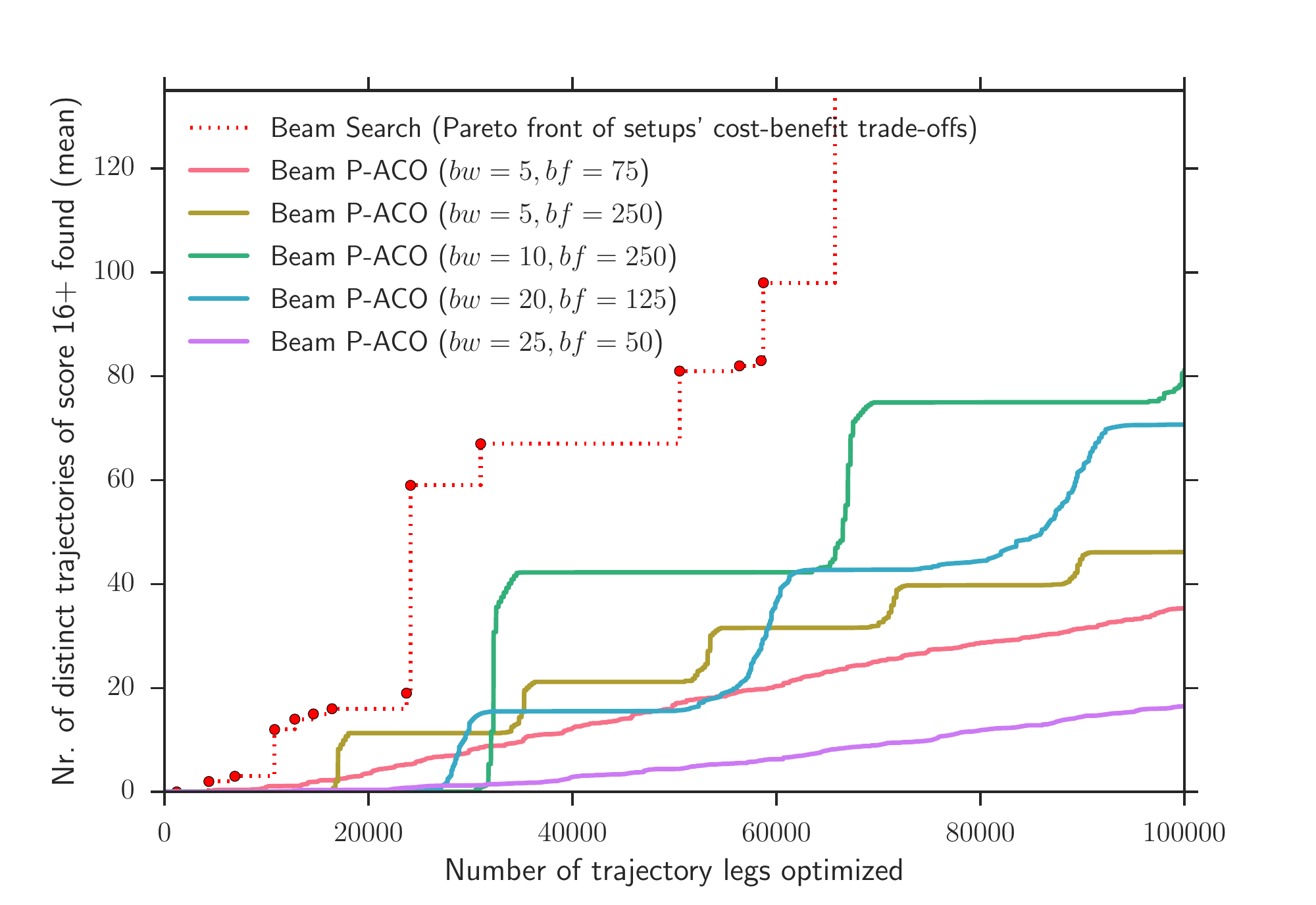}}
\caption{Quantity of top solutions: cumulative number of \textit{distinct} trajectories of score 16 or 17 found over time. Note the change of scale in the vertical axis.}
\label{fig:cumul_nr_seq}
%\end{figure*}

\vspace*{\floatsep}

%\begin{figure*}[t]%[!pt]
%\vspace{-10pt}
\centering
\subfloat[Beam Search]{\includegraphics[width=.33\textwidth,viewport = 22 8 537 363,clip]{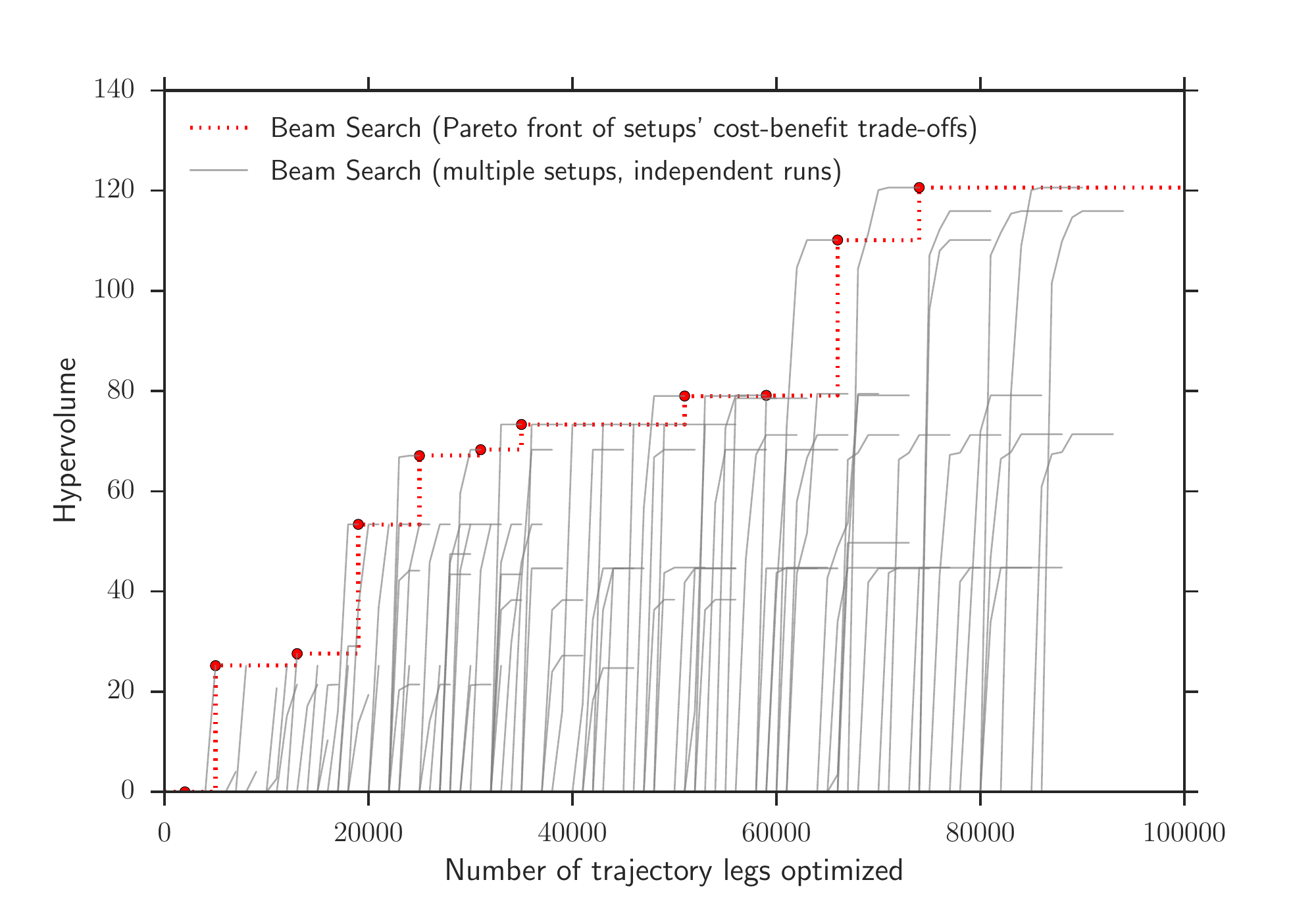}}
\hfil
\subfloat[Stochastic Beam]{\includegraphics[width=.33\textwidth,viewport = 22 8 537 363,clip]{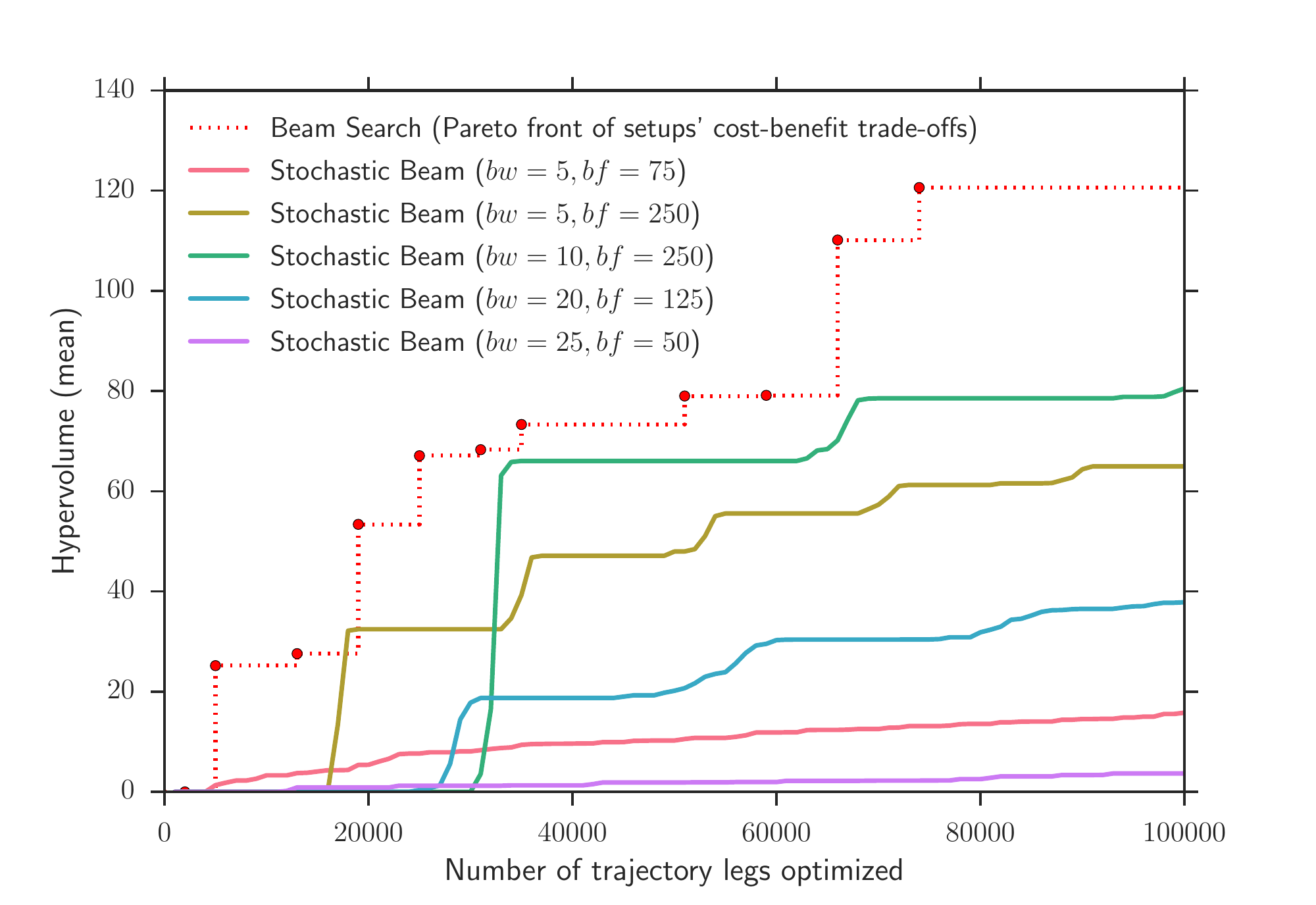}}
\hfil
\subfloat[Beam P-ACO]{\includegraphics[width=.33\textwidth,viewport = 22 8 537 363,clip]{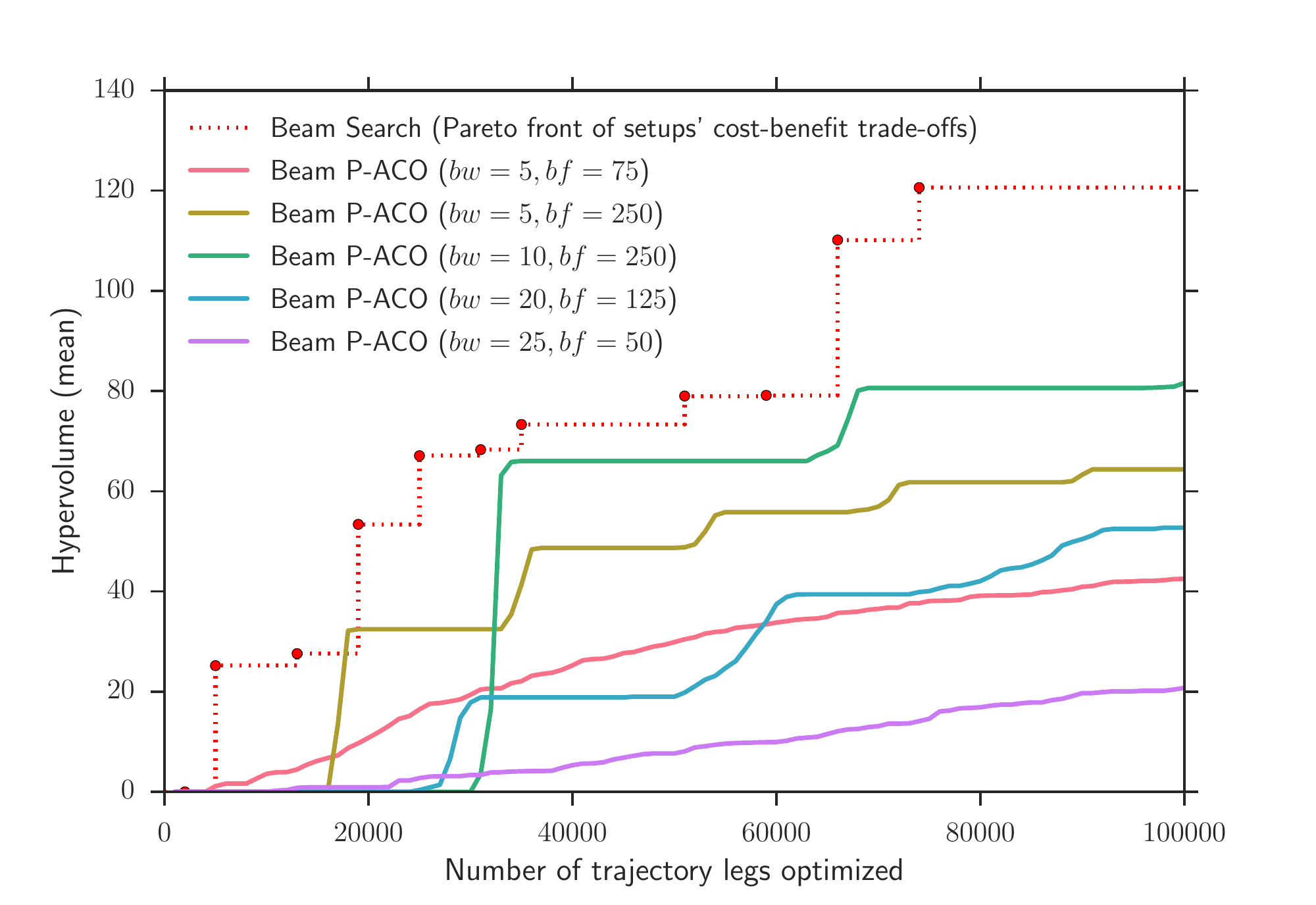}}
\caption{Quality of top solutions: growth of the dominated area of objective space over time (hypervolume of the Pareto front of score 16 trajectories found so far).}
\label{fig:hypervolume}
\end{figure*}

\begin{figure*}[t]%[!pt]
%\vspace{-10pt}
\centering
\subfloat[Stochastic Beam]{\includegraphics[width=.33\textwidth,viewport = 18 6 521 363,clip]{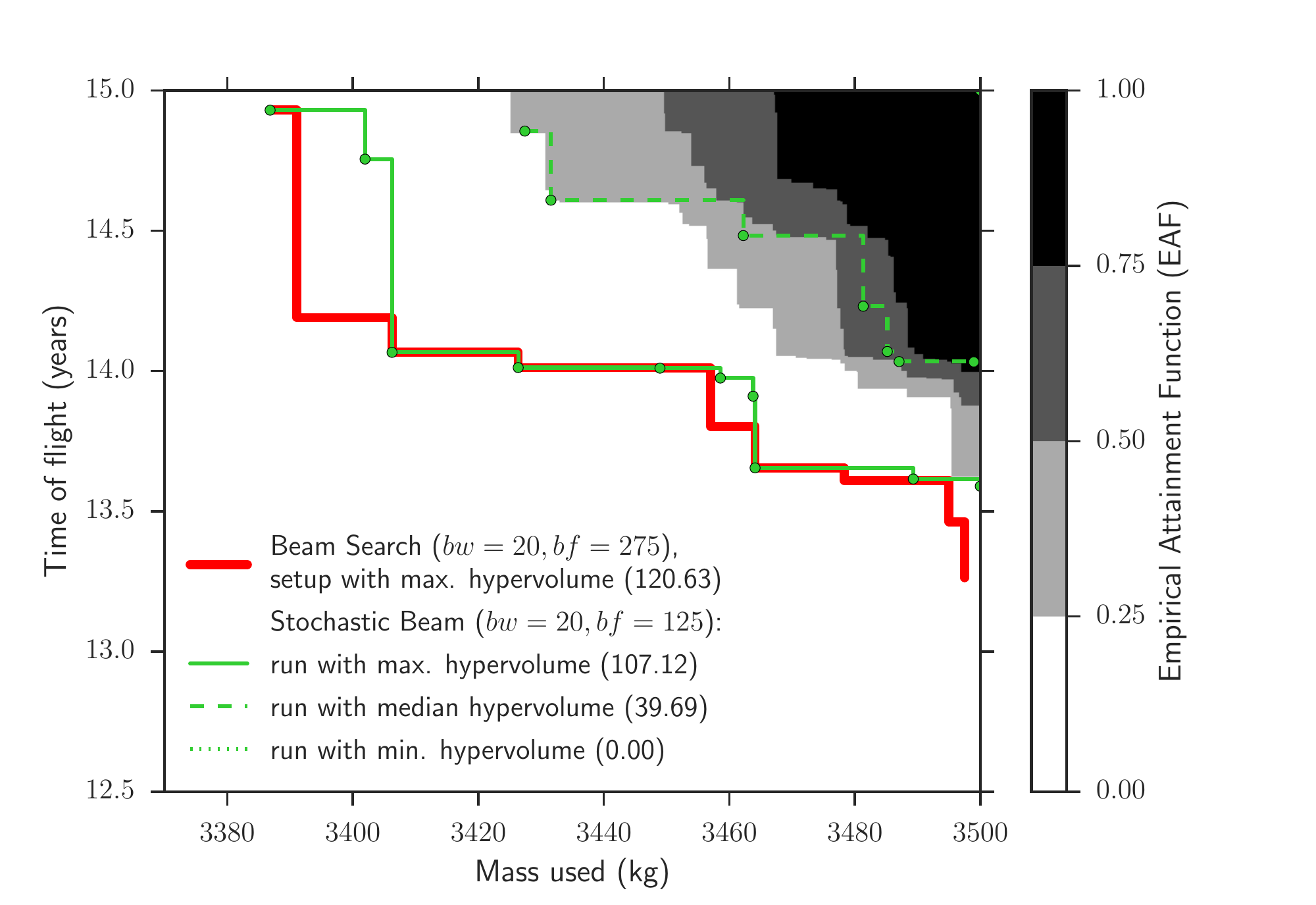}}
\hfil
\subfloat[Beam P-ACO]{\includegraphics[width=.33\textwidth,viewport = 18 6 521 363,clip]{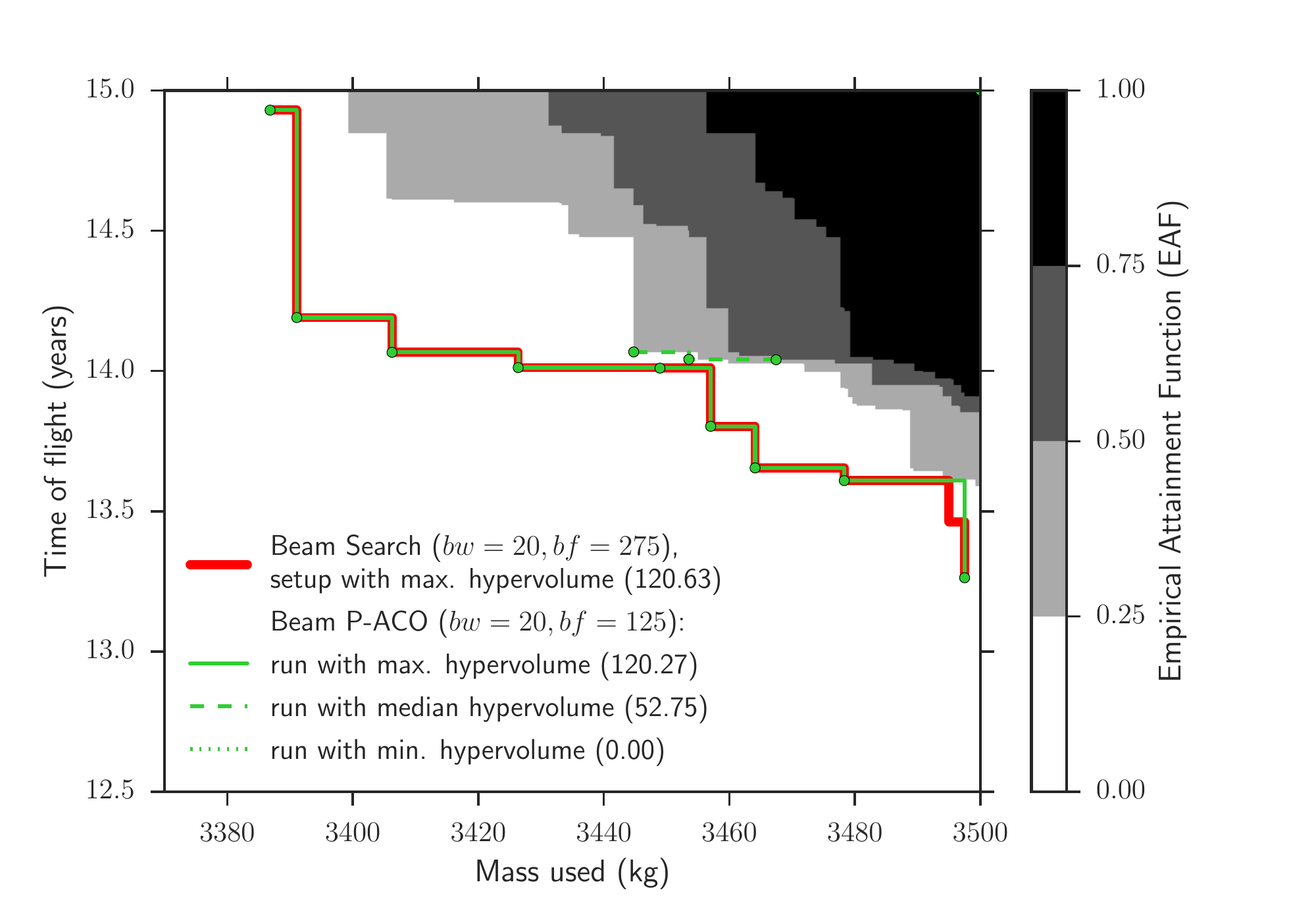}}
\hfil
\subfloat[Performance difference]{\includegraphics[width=.33\textwidth,viewport = 18 6 521 363,clip]{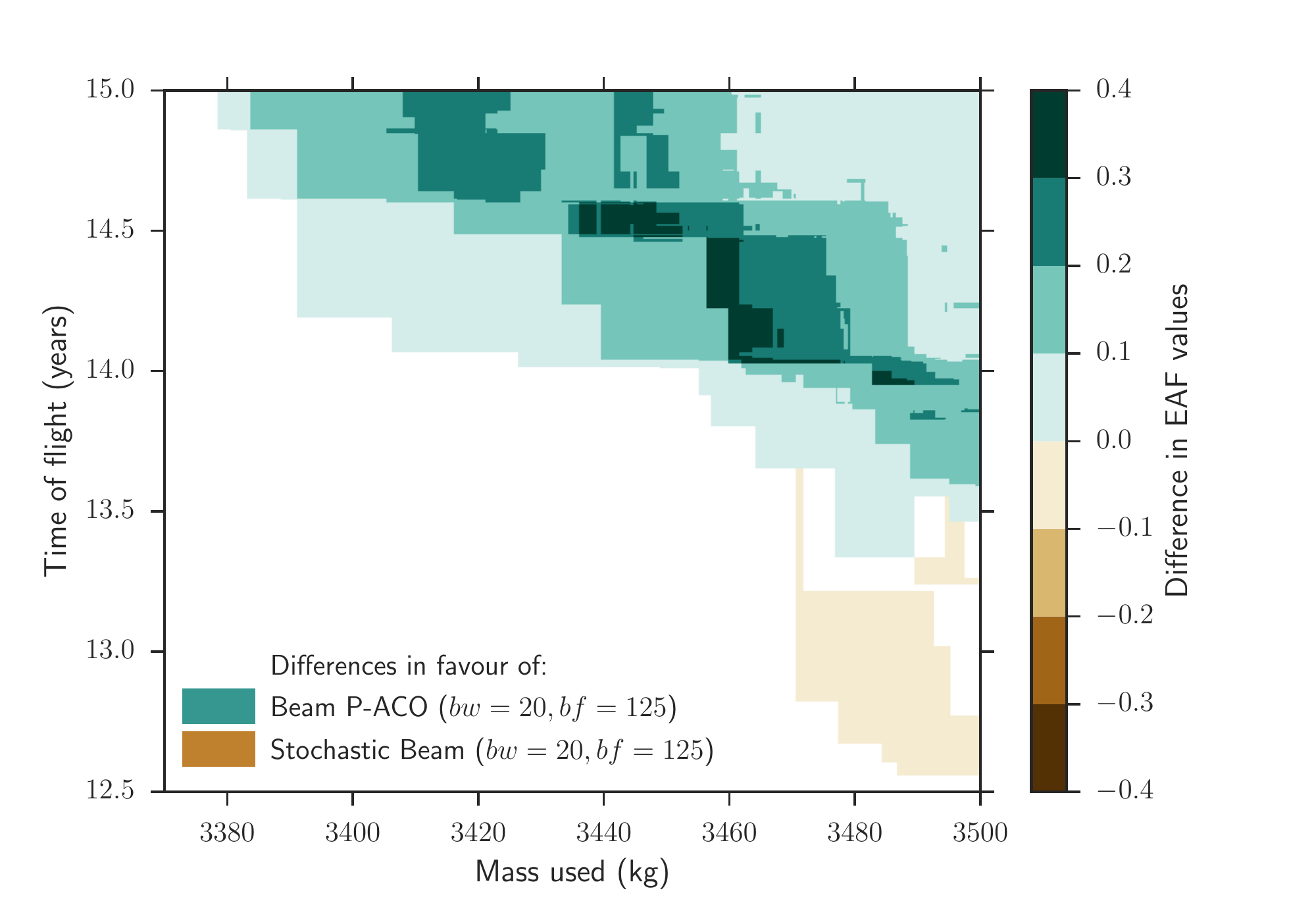}}
\caption{Empirical Attainment Functions: probabilities of objective space vectors being dominated (or matched) in an algorithm's run.
The darker a region is, the likelier it is that in a run a solution will be found that dominates it.
Considers the Pareto fronts of score 16 trajectories found up to the 100000 optimized legs threshold.}
\label{fig:eaf}
\end{figure*}

\subsection{Results}

The results obtained in the experimental evaluation are shown in Figures \ref{fig:param_sweep}--\ref{fig:eaf}.
Analyses into the quality of solutions found in a run consider their $h_s$ evaluations, in particular, the extent to which they minimize mass and time costs.
Though missions of score 17 were found in these experiments, they were only rarely found, and furthermore only a single distinct mission was ever found with that score. Therefore, analyses into the quality of solutions found consider exclusively the score 16 missions found in runs.
Specifically, we evaluate the Pareto fronts of score 16 missions found, and measure their coverage of the objective space using the hypervolume indicator \cite{Zitzler:03:MOs} -- a measure of the total area of objective space dominated by points in the Pareto front.
A reference point of 3500 kg and 15 years is used in all hypervolume calculations, corresponding to the limits set forth in the problem specification.
Runs that do not find any score 16 mission, and therefore have an empty Pareto front, have a hypervolume of 0.0.

\textbf{Parameter sweep of Beam Search configurations:}
% "how well does our baseline approach perform?"
%
the results from these experiments are shown in Fig. \ref{fig:param_sweep}.
Fig. \ref{fig:param_sweep}(a) shows the highest score reached among the missions designed in each run.
Fig. \ref{fig:param_sweep}(b) shows the hypervolume of the Pareto front of mass and time costs in score 16 missions found in each run.
Fig. \ref{fig:param_sweep}(c) shows the prevalence of unfeasible solutions in the continuous search spaces of asteroid transfers.
In the setups that reached score 17, for instance, on average only $\approx 23\%$ of the attempted asteroid transfers had a feasible solution.

\textbf{Quantity of top solutions:}
% "quantity"
%
Fig. \ref{fig:cumul_nr_seq} shows the results from the evaluation of the quantity of score 16 or 17 solutions found by each algorithm over time.
Only distinct trajectories count here to a run's totals -- two trajectories are equal if their asteroid sequence is exactly the same.
The three plots shown correspond, from left to right, to the results from Beam Search, Stochastic Beam, and Beam P-ACO. Fig. \ref{fig:cumul_nr_seq}(a) shows one curve for each of Beam Search's 118 evaluated setups. The algorithm being deterministic, each curve depicts also one single run.
Overlaid in this plot is the Pareto front of cost-benefit trade-offs attainable through different parameter settings: it shows the minimum number of trajectory legs that need to be optimized to obtain different amounts of top scoring trajectories.
This Pareto front is replicated in the other two plots to allow a performance comparison between the deterministic and randomized Beam Search variants.
The ``hooks'' shape seen, especially in Fig. \ref{fig:cumul_nr_seq}(a), results from the way Beam Search operates: most of its execution time is spent gradually descending through the tree, level by level. Eventually, the search reaches depth 15, at which point each branching event possibly leads to a score 16 solution being found. Hence, the sudden explosion in the total count. The plots for Stochastic Beam and Beam P-ACO also display this effect, but in them values are averaged over 100 runs, and consecutive tree searches (generations)
are chained together.

\textbf{Quality of top solutions:}
% "quality"
%
Figures \ref{fig:hypervolume} and \ref{fig:eaf} show the results from the evaluation of the quality of score 16 solutions found by each algorithm over time.
Fig. \ref{fig:hypervolume} is structured in the same way as Fig. \ref{fig:cumul_nr_seq}, so the description made above for it also applies here. Fig. \ref{fig:hypervolume} shows how the total dominated area of objective space (hypervolume) grows over time, as new score 16 trajectories are found and enter the Pareto front.
Fig. \ref{fig:eaf} takes a closer look at the setup $bw=20, bf=125$, over which both randomized Beam Search algorithms are seen in Fig. \ref{fig:hypervolume} reaching median performance (out of the five evaluated setups).
The plots shown are Empirical Attainment Functions (EAFs) \cite{GrunertdaFonseca2001}, which depict the likelihood of objective space vectors being dominated (or matched) in an algorithm's run. It aggregates into one visualization the final Pareto fronts of score 16 trajectories found across all 100 runs.
In Fig. \ref{fig:eaf}(a--b), the boundaries of the shaded areas show the 0.25, 0.5 and 0.75 ``attainment surfaces''. In other words, the areas having at least 25, 50 or 75$\%$ chance of being attained (dominated or matched) by points in a run's final Pareto front. Fig. \ref{fig:eaf}(c) takes the difference between the EAFs for Beam P-ACO and Stochastic Beam, and shows how they compare in terms of likelihood to attain different areas of objective space,
making clear the extent to which Beam P-ACO outperforms Stochastic Beam.
The Pareto front with greatest hypervolume found by deterministic Beam Search is shown for reference.

\section{Analysis \& discussion}
\label{sec:discuss}

Conceptually, we showed in this research a formal equivalence between four combinatorial optimization algorithms: Beam Search, Stochastic Beam, Beam P-ACO, and P-ACO.
We demonstrated how they can all be implemented in the same algorithm, and made accessible through minor parameter changes.

A surprising result, that validates the introduced phasing indicator, as well as the baseline multi-objective Beam Search algorithm employed here, was the discovery of a score 17 mission (17 asteroids visited, and fully investigated). Searches over this model of the GTOC5 problem, using these same initial conditions, had previously only reached a maximum score of 16 \cite{GTOC5:Izzo}. Furthermore, those searches, employing a Branch \& Prune tree search algorithm, took days to complete during the GTOC5 competition. The Beam Search variants under consideration could all find that single score 17 mission in runs lasting 10 to 20 minutes. Pure P-ACO is the exception here, having never surpassed a score 15 in our experiments (with a $bf=1$, the high chance of an asteroid transfer's optimization problem not having a feasible solution greatly limits performance).

The first research question in Sec. \ref{sec:intro} called for a demonstration of examples where randomized Beam Search variants would outperform the deterministic approach.
If we evaluate in terms of mean performance, then, as we can see from Figures \ref{fig:cumul_nr_seq}--\ref{fig:hypervolume}, the current experimental evaluation could not find any such cases.
At all computational cost thresholds we can find deterministic setups outperforming both of the randomized Beam Search variants, both in the quantity of score 16 solutions found, and in their quality. %
The deterministic algorithm was tuned to a considerably greater extent than the randomized ones (118 setups, against only 5), so it is possible that we are presenting a skewed view of each algorithm's capabilities. Alternatively, it may be that the specific problem we consider here is so resource-constrained, that the construction of long asteroid sequences actually demands for greedy branching decisions to be taken at every single step. In such a case, the search will not benefit from the greater tolerance for local sub-optimality present in the randomized Beam Search variants.

The second research question in Sec. \ref{sec:intro} considers how the addition of feedback (pheromones) in Beam P-ACO changes performance, by comparison with Stochastic Beam, where search proceeds along multiple independent generations with no feedback between them. Figures \ref{fig:cumul_nr_seq}--\ref{fig:eaf} show a clear positive effect of feedback on performance, both in the quantity and quality of top solutions found. This effect is seen to be larger the smaller the branching factor is. In other words, the greater the branching factor, the likelier it is for good solutions to be identified in a single generation, and thus the lower the benefit from feedback.

An identical comparison between randomized Beam Search variants, with and without pheromone updates, can be found in \cite[Sec. 6.2]{blum2005beamaco} (for a different hybrid algorithm, and different problems). There, pheromones are also found to improve performance, though in small amounts.
The current research goes beyond the analysis in \cite{blum2005beamaco} by uncovering the inverse relationship between branching factor and benefit from pheromones, and in demonstrating the superior performance of deterministic Beam Search over the randomized variants, given proper tuning.
% -------
% Quotes from \cite[Sec. 6.2]{blum2005beamaco}, page 1581:
% "The results are shown in Table 2(a). The First observation is that although the two versions with pheromone update are always better or equal to the two versions without pheromone update, the differences between all four versions are quite small."
% Again, the results show that the algorithm versions that use pheromone update are [...] slightly better than the other two versions."
% -------

Overall, an apt description of the behaviours displayed by the randomized algorithms is that they approximate the performance of deterministic Beam Search setups that use larger beam widths and branching factors -- probabilistic branching effectively picks a subset of the successor nodes that deterministic Beam Search with larger branching factors would follow.
The eventual success or failure of a run will then depend on how well those decisions align with those a better informed algorithm would take.
This can be seen on display in Fig. \ref{fig:eaf}. There, we see that in the extreme a randomized Beam Search run can closely approximate the Pareto front of score 16 trajectories found by an ``optimally" tuned deterministic Beam Search (with a $bf=125$, less than half of the $bf=275$ needed in the deterministic setting). However, it can also fail to find a single score 16 trajectory: 7 of the 100 runs in Fig. \ref{fig:eaf}(a), and 4 in Fig. \ref{fig:eaf}(b) could only reach a score 15, thus ending in this analysis with a min. hypervolume of 0.0.
Overall, the median hypervolume of 52.75 reached by Beam P-ACO in Fig. \ref{fig:eaf}(b) is higher than that reached in $\approx 60\%$ of the 118 deterministic Beam Search setups, and also higher than that reached in $50\%$ of the 32 deterministic setups that had equal or larger beam width and branching factors.

In a real setting, in the preliminary design phase, missions would be constructed not from a single set of initial conditions, as done here, but from a great number of them, possibly numbering in the thousands.
Being anytime algorithms, the randomized Beam Search variants investigated here can be employed in a racing approach \cite{Maron1997}, with multiple tree searches being executed in parallel. Computational effort would then be dynamically allocated across searches, as a function of the growing statistical evidence as to which searches lead to better missions.
In such a setting, the randomized Beam search variants would be preferable choices, over the deterministic algorithm.

\section{Conclusion}
\label{sec:conclusion}

We considered here a hard real-world problem of spacecraft trajectory design, featuring an interplay of combinatorial and (constrained) continuous optimization sub-problems, dealing at different levels with uncertain and multi-objective quality functions.
In this challenging domain, we investigated a number of extensions to Beam Search, the traditionally used approach to solve such problems. We provided an improved orbital phasing indicator, and its transformation into a probability distribution over candidate bodies to extend missions with. We then hybridized the search process with the well known Ant Colony Optimization algorithm, and investigated the behaviours of the resulting randomized Beam Search variants. We found them to have lower sensitivity to the beam width and branching factor parameter settings, while offering in each generation a partially-informed approximation to the behaviours of deterministic setups running at higher computational costs.

\section*{Acknowledgment}

Lu{\'i}s F. Sim{\~o}es was supported by FCT (Minist\'erio da Ci\^encia e Tecnologia) Fellowship SFRH/BD/84381/2012.

\bibliographystyle{splncs03}
\bibliography{simoes}

% that's all folks
\end{document}